\documentclass{article}

\usepackage[preprint]{neurips_2024}

\usepackage{caption}
\usepackage{subcaption}
\usepackage{graphicx}
\usepackage[utf8]{inputenc} % allow utf-8 input
\usepackage[T1]{fontenc}    % use 8-bit T1 fonts
\usepackage{hyperref}       % hyperlinks
\usepackage{url}            % simple URL typesetting
\usepackage{booktabs}       % professional-quality tables
\usepackage{amsfonts}       % blackboard math symbols
\usepackage{nicefrac}       % compact symbols for 1/2, etc.
\usepackage{microtype}      % microtypography
\usepackage{xcolor}         % colors
\usepackage{amsmath}

\title{%1-Data Leakage from Convolutional Layer Gradients: An Analytical Approach\\
 R-CONV: An Analytical Approach for Efficient Data Reconstruction via Convolutional Gradients}

\author{%
Tamer Ahmed Eltaras\\
Control and Computer Engineering Dept\\
  Politecnico di Torino\\
  Italy, Turin 10129 \\
  \texttt{tamer.ahmedeltaras@polito.it} 
  \And
  Qutaibah Malluhi\\
 Computer Science and Engineering Dept\\
  Qatar University\\
  Qatar, Doha 2713 \\
  \texttt{qmalluhi@qu.edu.qa} 
  \And
 Alessandro Savino\\
  Control and Computer Engineering Dept\\
  Politecnico di Torino\\
  Italy, Turin 10129 \\
  \texttt{alessandro.savino@polito.it} 
  \And
  Stefano Di Carlo\\
  Control and Computer Engineering Dept\\
  Politecnico di Torino\\
  Italy, Turin 10129 \\
  \texttt{stefano.dicarlo@polito.it}
  \And
  Adnan Qayyum\\
  Computer Science Information\\
  Technology University Dept\\
  Punjab, Pakistan \\
  \texttt{adnan.qayyum@itu.edu.pk}
  \And
  Junaid Qadir\\
  Computer Science and Engineering Dept\\
  Qatar University\\
  Qatar, Doha 2713 \\
  \texttt{jqadir@qu.edu.qa} 
}

\begin{document}

\maketitle

\begin{abstract}
\label{abstract}
In the effort to learn from extensive collections of distributed data, federated learning has emerged as a promising approach for preserving privacy by using a gradient-sharing mechanism instead of exchanging raw data. However, recent studies show that private training data can be leaked through many gradient attacks. While previous analytical-based attacks have successfully reconstructed input data from fully connected layers, their effectiveness diminishes when applied to convolutional layers. This paper introduces an advanced data leakage method to efficiently exploit convolutional layers' gradients. We present a surprising finding: even with non-fully invertible activation functions, such as ReLU, we can analytically reconstruct training samples from the gradients. To the best of our knowledge, this is the first analytical approach that successfully reconstructs convolutional layer inputs directly from the gradients, bypassing the need to reconstruct layers' outputs. Prior research has mainly concentrated on the weight constraints of convolution layers, overlooking the significance of gradient constraints. Our findings demonstrate that existing analytical methods used to estimate the risk of gradient attacks lack accuracy. In some layers, attacks can be launched with less than 5 \% of the reported constraints. The source code is available for download from:
\begin{center}
  \url{https://github.com/tamertaras91/RR-CONV.git}
\end{center}
\end{abstract}

\section{Introduction}
\label{introduction}

As the demand for data increases, the need to efficiently leverage vast datasets becomes increasingly urgent. In recent years, Federated Learning (FL) \citep{mcmahan2017communication} has emerged as a promising paradigm for addressing the conflict between learning from large sets of distributed data and fulfilling privacy demands. FL enables collaborative model training among multiple clients without exposing their raw data. In FL, clients work together to train a global model under the coordination of a central server across multiple rounds. During each round, clients train the model locally using their private training data and then transmit gradients to the server for aggregation and global update. Recently, it has been demonstrated that this approach is susceptible to attacks that can recover the training data from the gradient information exchanges, which raises questions about its suitability as a privacy-preserving distributed machine-learning paradigm~\citep{zhao2020idlg,zhu2019deep,wang2019beyond,wei2020framework,geiping2020inverting}. 

Gradient attacks are highly effective attacks that recover private training data from gradient vectors. These attacks cast significant privacy challenges on distributed learning from clients with sensitive data, where clients must share gradients. Defending against such attacks requires understanding when and how privacy leakage happens. In this work, we propose a novel closed-form attack algorithm that significantly enhances the capability to extract training data from the gradient information of neural networks, with a particular emphasis on convolution layers. Previous work that studied the gradient leakage from convolution layers \citep{zhu2020r,chen2021understanding} have limited their methods to scenarios where only fully-invertible activation functions are employed and mainly relied on weight constraints of the convolution layers to formulate the equations used in their approach. 

In our work, we overcome these limitations and propose a method that efficiently tackles non-fully invertible functions like ReLU. In contrast to \citet{zhu2020r}, who found that convolution neural networks (CNNs) have many more weight constraints than gradient constraints, and that the valid gradient constraints are much fewer in CNNs compared to fully connected layers, our findings reveal the crucial importance of gradient constraints in reconstructing input data in convolution layers. This significant factor has been previously underestimated in the literature. We advance the understanding of gradient leakage by demonstrating the substantial value of gradient constraints. Our method leverages these constraints to reconstruct training data more effectively, especially in layers where the feature map channels increase. Key contributions of our research include:
\begin{itemize}
    \item Proposing, R-CONV, an advanced analytical method that successfully reconstructs training data from the gradients of convolutional layers, even when non-fully invertible activation functions such as ReLU are used.
    \item Highlighting the significance of exploiting gradient constraints, which, in some layers, prove to be more revealing than weight constraints.
    \item Exploring how variations in convolutional layer parameters, including kernel size, stride, and padding, influence the effectiveness of gradient-based attack techniques.
    \item Demonstrating that existing analytical methods used for risk estimation of gradient attacks lack accuracy, as they underestimate the importance of gradient constraints in convolution layers.
\end{itemize}

\section{Related Work}
\label{sec:related}
% \subsection{Gradient Leakage in FL}
% Over the past few years, various studies have demonstrated the feasibility of inferring privacy-related attributes from the shared model updates in FL. 
Recent studies have shown that shared model updates in FL can be exploited to infer privacy-related attributes. Conventional ``shallow'' gradient leakage attacks focused on inferring membership properties of subject samples \citep{melis2019exploiting} and generating such samples that look similar to those from training data using generative models \citep{hitaj2017deep}. Building upon these techniques, \cite{zhu2019deep} introduced the idea of Deep Gradient Leakage (DLG) and demonstrated that shared model gradients can be exploited to fully reconstruct the training data. This has attracted significant research interest resulting in the development of various methods for reconstructing input training samples, which can be categorized into optimization-based and analytical approaches. 

% Since then, this line of research has attracted significant attention, resulting in the development of various methods for reconstructing input training samples. These methods can be broadly classified into two major categories: optimization-based and analytical approaches, described below.

\textbf{Optimization-Based Approaches:} \cite{zhu2019deep} proposed an optimization-based algorithm that takes the random ``dummy'' gradients and corresponding class labels to process them through the forward and backward passes. Instead of optimizing model weights, they proposed minimizing the distance between the dummy gradients and the real gradients, resulting in the reconstruction of the original input. The authors demonstrated that accurate pixel-level reconstruction is practically feasible for a maximum batch size of 8. However, their proposed technique is only compatible with simple neural network architectures having sigmoid activation and only works for low-dimensional CIFAR images. Consequently, the proposed method cannot reconstruct high-dimensional images and complex neural networks with ReLU activations. \cite{geiping2020inverting} followed the same problem formulation as that of \cite{zhu2019deep} and introduced an extra regularization term in the optimization program that provided significant performance gain, enabling the reconstruction of high-dimensional images. Specifically, they used cosine similarity for measuring reconstruction loss (instead of Euclidean distance) and total variation as the regularizer. Their experimental results demonstrate that the proposed technique could reconstruct low-dimensional images for a maximum batch size of 100. In a similar study, \cite{zhao2020idlg} have attempted to improve DLG by consistently inferring reference labels from gradients. Similarly, to increase the performance of gradient inversion attacks and improve their performance under realistic settings, the use of other techniques such as auxiliary information \citep{yin2021see}, and different regularization techniques (such as generative image priors \citep{jeon2021gradient}) have been explored in the literature. Moreover, to quantify the impact of hyperparameter choices on the attack performance, \cite{wei2020framework} presented a framework for gradient leakage attacks in FL. 

% To protect against gradient leakage attacks, it is crucial to understand underlying causes, especially when we are dealing with black-box models. 

\textbf{Analytical Approaches:} Literature has extensively delved into providing theoretical insights into the occurrence of gradient inversion. For instance, \cite{zhu2020r} presented a closed-form iterative method to reconstruct training data from the model's gradients. In addition, to assess the susceptibility of certain network architectures to gradient attacks, the authors presented a rank analysis technique that is agnostic to whether the attack procedure is optimization-based or closed-form. In a similar study, \cite{chen2021understanding} presented a Combined Optimization Attack (COPA) that provides general insights into the mechanism of training data leakage through a more informative formulation of the objective function that makes it clearer to identify the source of constraints. Recently, \cite{zhang2024understanding} proposed an analytical tool that uses the Inversion Influence Function (I2F) to analyze the relationship between gradient inversion and perturbation. I2F efficiently scales for deep networks and requires only oracle access to gradients and Jacobian-vector products. They offered key insights, including highlighting the importance of non-homogeneous gradient perturbation, the impact of Jacobian structures on privacy protection, and the role of model parameter initialization in shaping privacy risks. They argued that these findings can be leveraged to develop effective defense mechanisms against gradient leakage attacks. Our proposed approach resembles the aforementioned analytical methods that are also focused on evaluating privacy risks in convolutional layers of CNN architectures. \cite{wang2023reconstructing} showed that it is possible to fully reconstruct training data using a single gradient query, even if the model is not efficiently trained. Most of the aforementioned analytical gradient inversion attacks assume that activation functions used in the underlying model are fully invertible. Consequently, they will not work for models using ReLU-based activation functions. Our approach overcomes these limitations, with theoretical and empirical evidence demonstrating its effectiveness.

\section{Methodology}
\label{sec:Methodlogy}
In this section, we outline our methodology for reconstructing training examples by exploiting the gradients $ \frac{\partial \ell}{\partial W} $ and $  \frac{\partial \ell}{\partial b}$, along with initial parameters $W$ and $b$. Initially, we consider a simplified model: a CNN with a single convolution layer followed by a fully connected layer. The activation function $A$ is applied between these layers. Although our methodology initially focuses on this simple architecture, it is generalizable to more complex multi-layer networks. We structure our approach into three key phases, each targeting a different network architecture component.

\subsection{Gradient Computation and Input Reconstruction from Fully Connected Layer}

Consider the input vector to the fully connected layer as $X=(x_1,x_2,\cdots,x_i)$, and the network's loss function expressed as:
\[\ell=\ell(f(X),y)\]
The gradient vector w.r.t. the layer's input is $\frac{\partial\ell}{\partial X}=(\frac{\partial\ell}{\partial X_1},\frac{\partial\ell}{\partial X_2},\cdots,\frac{\partial\ell}{\partial X_i})$. Assume $Z=(z_1,z_2,\cdots,z_c)$ is the output of the layer, where $c$ denotes the number of nodes (classes), and $b=(b_1,b_2,\cdots,b_c)$ represents the bias vector. The gradient vector w.r.t. the biases is $\frac{\partial\ell}{\partial b}=(\frac{\partial\ell}{\partial b_1},\frac{\partial\ell}{\partial b_2},\cdots,\frac{\partial\ell}{\partial b_c})$, and the weight vector is $W=(w_{11},w_{12},\cdots,w_{21},\cdots,w_{ic})$, with each $w_{ic}$ denoting the weigh between input $x_i$ and node $c$. The gradient vector w.r.t. the weights is $\frac{\partial\ell}{\partial W}=(\frac{\partial\ell}{\partial w_{11}},\frac{\partial\ell}{\partial w_{12}},\cdots,\frac{\partial\ell}{\partial w_{ic}})$. The output of the layer can be expressed as:
\[Z = WX + b,\]
And the output for a node $m$ can be represented as:

   \[ z_m = \sum_{i=1}^{i} (x_i w_{im}) + b_m\]

So we can compute for an input $x_n$ :
\begin{equation}
    \frac{\partial z_m}{\partial b_m}=1, \frac{\partial z_m}{\partial x_n} = w_{nm}
\end{equation}

Based on the chain rule we can write:
   \begin{equation}
\frac{\partial \ell}{\partial b_m}=\frac{\partial \ell}{\partial z_m} \times \frac{\partial z_m}{\partial b_m}
   \end{equation}     
   \begin{equation}
\frac{\partial \ell}{\partial x_n}=\frac{\partial \ell}{\partial z_m} \times \frac{\partial z_m}{\partial x_n}
   \end{equation} 
By substituting from equation $1$ into equation $2$ we can write,
\begin{equation}
    \frac{\partial \ell}{\partial z_m}=\frac{\partial \ell}{\partial b_m}
\end{equation}
and by substituting from equation (1) and (4) in (3) we can get:

\begin{equation}
    \frac{\partial \ell}{\partial x_n}=\frac{\partial \ell}{\partial b_m} \times w_{nm}
\end{equation}
Therefore, we can uniquely compute the above gradient as we have access to the weight and bias gradients, and the initial parameters.\\
Equation $(5)$ represents the gradient w.r.t. one node. To complete the total gradient w.r.t. the input $x_n$:

can be derived from gradients w.r.t. W and gradients w.r.t. b, i.e.

\begin{equation}
    \frac{\partial \ell}{\partial x_n}= \sum_{c=1}^{c}\frac{\partial \ell}{\partial x_n}
\end{equation}

Now, we need to propagate the gradient vector w.r.t. input to the convolution layer. To do that, we need to construct the input vector $X_i$. In their work, \citet{aono2017privacy} showed that the input $X$ can be derived from gradients w.r.t. $W$ and gradients w.r.t. $b$. Following the equations above and to construct for an input $n$:

     \[ \frac{\partial\ell}{\partial w_{nm}} = \frac{\partial\ell}{\partial z_m}\times\frac{\partial z_m}{\partial w_{nm}}\\, \frac{\partial z_m}{\partial w_{nm}}=x_n\]
          \[ \frac{\partial\ell}{\partial w_{nm}} = \frac{\partial\ell}{\partial z_m} x_n\]
         \[x_n= \frac{\partial\ell}{\partial w_{nm}} / \frac{\partial\ell}{\partial z_m} \]
and by substituting from equation $(4)$:
\begin{equation}
    x_n= \frac{\partial\ell}{\partial w_{nm}} / \frac{\partial\ell}{\partial b_m}
\end{equation}
As long as there exists at least one node $m$ with $\frac{\partial \ell}{\partial b_m}\neq 0$, the single input $x_n$ is recovered perfectly.\\

So from equations $(6)$ and $(7)$ we can fully recover both the gradient w.r.t. the input $\frac{\partial\ell}{\partial X}$  and the input vector $X$ analytically from the gradients w.r.t. $W$, gradients w.r.t. $b$, and initial parameters $W$ and $b$.

\subsection{Propagating Gradient Through Activation Function}
\label{propagating}
To drive the gradient vector w.r.t. the output of the convolution layer $\frac{\partial\ell}{\partial O}=(\frac{\partial\ell}{\partial o_1},\frac{\partial\ell}{\partial o_2},\cdots,\frac{\partial\ell}{\partial o_i})$, we employ the chain rule to propagate the gradient vector w.r.t. to the input of the fully connected layer through the activation function.

Let $A$ denote the activation function, where $O$ represents the input to the activation function from the convolution layer, and $X$ denotes the output fed into the fully connected layer $X= A(O)$. The gradient vector w.r.t. the Output $O$ is computed as the element-wise product of the derivative of the activation function $A^{\prime}(O)$ and the gradient vector w.r.t. the input of the fully connected layer $\frac{\partial\ell}{\partial X}$:
\[ \frac{\partial\ell}{\partial O} = \frac{\partial\ell}{\partial X}\times \frac{\partial X}{\partial O}\]
\begin{equation}
    \frac{\partial\ell}{\partial O} =  \frac{\partial\ell}{\partial X}\times A^{\prime}(O) 
\end{equation}
Considering that the derivatives of the most commonly used activation functions in neural networks can be expressed solely in terms of the function's output, we can efficiently compute this derivative using the constructed input vector of the fully connected layer. Table \ref{Activation_function-table} summarizes the derivatives of various activation functions commonly employed in neural networks, expressing them in terms of the output of the activation function $X$.
%Our approach is vaild as long as the activation function exhibit this property.

%For example of we supposed that the activation function is ReLU so the derivative is as follow:
%\[
 %A’(O) = 
  %   \begin{cases}
   %  1 & \mbox{ if $ x > 0 $}\\
    % 0 & \mbox{ if $ x \leq 0 $}
     %\end{cases}   
%\]

%So we can write:
%\begin{equation}
 %    \frac{\partial\ell}{\partial O} = 
  %   \begin{cases}
    %\frac{\partial\ell}{\partial X} & \mbox{ if $ x > 0 $}\\
   %0 & \mbox{ if $ x \leq 0 $}
    % \end{cases}  
%\end{equation}

\begin{table}
\scriptsize
  \caption{Derivative of activation functions.}
  \label{Activation_function-table}
  \centering
  \begin{tabular}{lll}
    \toprule

    Name     & Equation     & Derivative expressing in $X$ \\
    \midrule
    Sigmoid & $ A(O)=\frac{1}{1+e^-O}$ & $A^{\prime}(O)=X(1-X)$      \\
    Tanh & $ A(O)=\tanh{(O)} $ & $A^{\prime}(O)=1-X^2$      \\
    ArcTan & $ A(O)=\tan^{-1} {(O)}$ & $A^{\prime}(O)=\frac{1}{1-\tan(X)^2}$      \\
    SoftPlus & $ A(O)=\log_e {(1+e^O)}$ & $A^{\prime}(O)=\frac{1}{1+\ln(e^X-1)}$      \\
    ReLU     & $ A(O) = 
     \begin{cases}
     O & \mbox{ if $ O > 0 $}\\
     0 & \mbox{ if $  O \leq 0 $}
     \end{cases}  $  & $ A^{\prime}(O) = 
     \begin{cases}
     1 & \mbox{ if $ X > 0 $}\\
    
     0 & \mbox{ if $ X = 0 $}
     \end{cases}  $    \\
        Leaky ReLU     & $ A(O) = 
     \begin{cases}
      O & \mbox{ if $ O \geq 0 $}\\
      0.01 O & \mbox{ if $ O < 0 $}
     \end{cases}  $  & $ A^{\prime}(O) = 
     \begin{cases}
     1 & \mbox{ if $ X \geq 0 $}\\
    
     0.01 & \mbox{ if $ X < 0 $}
     \end{cases}  $    \\ 
    Parameteric ReLU     & $ A(O) = 
     \begin{cases}
      O & \mbox{ if $ O \geq 0 $}\\
     \alpha O & \mbox{ if $ O < 0 $}
     \end{cases}  $  & $ A^{\prime}(O) = 
     \begin{cases}
     1 & \mbox{ if $ X \geq 0 $}\\
    
     \alpha & \mbox{ if $ X < 0 $}
     \end{cases}  $    \\
     ELU     & $ A(O) = 
     \begin{cases}
      O & \mbox{ if $ O \geq 0 $}\\
     \alpha (e^O-1) & \mbox{ if $ O < 0 $}
     \end{cases}  $  & $ A^{\prime}(O) = 
     \begin{cases}
     1 & \mbox{ if $ X \geq 0 $}\\
    
     X+\alpha & \mbox{ if $ X < 0 $}
     \end{cases}  $    \\

    \bottomrule
  \end{tabular}
\end{table}

\subsection{Gradient Computation and Input Reconstruction in Convolution Layer}
In a typical CNN, the convolution operation involves sliding a kernel (or filter) across the input data and computing the dot product of the kernel and the corresponding parts of the input it covers. To illustrate the equations for deriving both the input and the gradient with respect to the input, let's consider a simple example with a $3x3$ input matrix and a $2x2$ filter kernel. In the forward pass, the output $O$ is expressed as follows: 
\[o_{1,1}=x_1w_{1,1}+x_2w_{1,2}+x_4w_{1,4}+x_5w_{1,5}\]
\[o_{1,2}=x_2w_{1,1}+x_3w_{1,2}+x_5w_{1,4}+x_6w_{1,5}\]
\[o_{1,3}=x_4w_{1,1}+x_5w_{1,2}+x_7w_{1,4}+x_8w_{1,5}\]
\begin{equation}
     o_{1,3}=x_4w_{1,1}+x_5w_{1,2}+x_7w_{1,4}+x_8w_{1,5}
\end{equation}
\subsubsection{Gradient Computation}
Applying the chain rule, the gradient w.r.t. $x_1$ is expressed as:
\[\frac{\partial \ell}{\partial x_1}=\frac{\partial \ell}{\partial o_{1,1}}\times\frac{\partial o_{1,1}}{\partial x_1}\]

And from previous equations we can obtain $\frac{\partial o_{1,1}}{\partial x_1}=w_{1,1}$.
Thus,
\[\frac{\partial \ell}{\partial x_1}=\frac{\partial \ell}{\partial o_{1,1}}\times w_{1,1}\]
Similarly, for $x_2$, as it contributes to both outputs $o_{1,1}$ and $o_{1,2}$ through weights $w_{1,2}$ and $w_{1,1}$ respectively:

\[\frac{\partial \ell}{\partial x_2}=\frac{\partial \ell}{\partial o_{1,1}}\times w_{1,2} + \frac{\partial \ell}{\partial o_{1,2}}\times w_{1,1}\]
In general, the gradient w.r.t. each input to the convolution layer can be derived by summing the dot product between each output the data point contributes to and the corresponding weight. If we assume that input $x_n$ contributes to the number of outputs $m$ with indices stored in list $v$, the input contributes to these outputs through the weights with indices stored in $t$, then the general formula can be written as follows:
\begin{equation}
        \frac{\partial \ell}{\partial x_n}=\sum_{m=1}^{m}\frac{\partial \ell}{\partial o}[v[m]]\times w[t[m]]
\end{equation}
\subsubsection{Input Reconstruction}

Applying the chain rule the gradient w.r.t. $w_{1,1}$ is expressed as:

\[\frac{\partial \ell}{\partial w_{1,1}}=\frac{\partial \ell}{\partial o_{1,1}}\times\frac{\partial o_{1,1}}{\partial w_{1,1}}\]

From equations at $(9)$ we can obtain: 
                        \[\frac{\partial o_{1,1}}{\partial w_{1,1}}=x_1,\frac{\partial o_{1,2}}{\partial w_{1,1}}=x_2, \frac{\partial o_{1,3}}{\partial w_{1,1}}=x_4, \frac{\partial o_{1,1}}{\partial w_{1,1}}=x_5 \]
Thus,
                \[\frac{\partial \ell}{\partial w_{1,1}}= \frac{\partial \ell}{\partial o_{1,1}}\times x_1 +\frac{\partial \ell}{\partial o_{1,2}}\times x_2+\frac{\partial \ell}{\partial o_{1,3}}\times x_4+\frac{\partial \ell}{\partial o_{1,1}}\times x_5 \]
Similarly, for other weights:
                \[\frac{\partial \ell}{\partial w_{1,2}}= \frac{\partial \ell}{\partial o_{1,1}}\times x_2 +\frac{\partial \ell}{\partial o_{1,2}}\times x_3+\frac{\partial \ell}{\partial o_{1,3}}\times x_5+\frac{\partial \ell}{\partial o_{1,1}}\times x_6 \]
                \[\frac{\partial \ell}{\partial w_{1,3}}= \frac{\partial \ell}{\partial o_{1,1}}\times x_4 +\frac{\partial \ell}{\partial o_{1,2}}\times x_5+\frac{\partial \ell}{\partial o_{1,3}}\times x_7+\frac{\partial \ell}{\partial o_{1,1}}\times x_8 \]
                \[\frac{\partial \ell}{\partial w_{1,4}}= \frac{\partial \ell}{\partial o_{1,1}}\times x_5 +\frac{\partial \ell}{\partial o_{1,2}}\times x_6+\frac{\partial \ell}{\partial o_{1,3}}\times x_8+\frac{\partial \ell}{\partial o_{1,1}}\times x_9 \]
Similarly, we can obtain an equal number of equations from gradients w.r.t. $w_2$ and $w_3$, which provides sufficient equations to reconstruct the input $X$.

In general, from each gradient w.r.t. to a weight in the convolution layer, we can obtain an equation involving some of the input data points. if we assume a specific weight $w_{d,i}$ that contributes to the $m$ data outputs through the input data points in list $r$, we can drive the general formula as follows:
\begin{equation}
    \frac{\partial \ell}{\partial w_{d,i}}= \frac{\partial \ell}{\partial o_{d,1}}\times x[r[1]] +\frac{\partial \ell}{\partial o_{d,2}}\times x[r[2]]+\dots+\frac{\partial \ell}{\partial o_{d,m}}\times x[r[m]] 
\end{equation}

With a suitable number of filters, we can establish a set of linear equations to reconstruct $X$.

\section{Analysis of Recursive Gradient on Convolutional Layers}
In this section, we provide an analytical approach to estimate the feasibility of performing an analytical recursive gradient attack on a convolutional layer. Although our methodology mainly relies on gradient constraints, to provide a full analysis, we will also consider the weight constraints and then later limit their significance depending on the activation functions that are employed. In the forward pass, the output $O^l$ of a convolutional layer $l$ is given by the convolution of the input data of this layer $X^l$ with the filter weights $W^l$ represented as $O^l=A^lX^l$, where $A^l$ is the matrix representing the convolution operation with weights. In the backward pass, the gradients w.r.t. the filter weights $\nabla W^l$ are given by the convolution operation between the input $X^l$ and the gradients w.r.t. the output $\nabla O^l$ represented as $\nabla W^l= B^lX^l$, where $B^l$ represents the convolution operations with the gradients. To solve for $X^l$, we combine the forward and backward pass equations into a single system:\[
\begin{bmatrix}
A^l \\
B^l
\end{bmatrix} X^l = \begin{bmatrix}
O^l \\
\nabla W^l
\end{bmatrix}
\]
To successfully reconstruct $X^l$, the rank of the matrix $\begin{bmatrix}
A^l \\
B^l
\end{bmatrix}$ must be equal to the number of unknowns $|X^l|$.
The number of equations in $A^l$ represents the number of weight constraints, and the number of equations in $B^l$ represents the number of gradient constraints.

\textbf{Weight constraints:} The number of weight constraints depends on the number of output data points $|O^l|$ that could be constructed from the subsequent layer $l+1$. The number of constraints depends on the type of activation function employed on the output of the layer. In the case the activation function is fully invertible, $|A^l|=|O^l|$, equating to the number of output data points; if non-invertible, $|A^l|= 0$; if partially invertible, $|A^l|$ is a subset of $|O^l|$, depending on how many data points we can reconstruct.

\textbf{Gradient constraints:} The number of gradient constraints, in case we are able to derive the gradients w.r.t. to the output of the layer $\nabla O^l$, is $|B^l|=|W^l|$, equal to the number of weights in the layer. The ability to derive $\nabla O^l$ depends on the success of propagating the gradients w.r.t. the subsequent layer input $\nabla X^{l+1}$ through the activation function, as demonstrated in the methodology (Section \ref{propagating}).

It is worth noting that as the number of weight constraints depends on the output dimension, and the number of gradient constraints depends on the number of weight in the first layer, assuming the activation function is invertible, the number of weight constraints will be greater than the number of gradient constraints $(|A^l| > |B^l|)$ because in the first layer, the output dimension is still large, and the number of input channels is few. However, as we move deeper into the convolutional layers, the output dimension typically decreases, and the number of channels increases, resulting in the number of gradient constraints becoming much larger than the number of weight constraints $(|A^l| < |B^l|)$. The previous work by~\cite{zhu2020r} has inaccurately reported that there are very few valid gradient constraints in a convolutional layer.
To mitigate the risk of gradient attacks, we must ensure that the following conditions hold: $|X^l > |A^l| +|B^l|$. Our findings indicate that the number of filters in the convolution layer is the key factor enabling the analytical attack. To quantify the risk using the number of filters, suppose the dimension for the input is $(H*H*N)$, the number of filters is $F$, and the kernel size is $K$, with stride $S$ and padding $P$:\[|A^l|= (\frac{H+2P-K}{S}+1) * F\] \[|B^l|= K^2 * F\]
Adjusting these parameters allows us to control the number of weight and gradient constraints, thus influencing the feasibility and risk of gradient attacks.
Another important point is that the above analysis also holds for the case of using ResNet (Residual Network). In our approach, we construct all the gradients with respect to the outputs of the layers. To reconstruct the gradient with respect to the input of the residual block, we simply combine the gradients from both the residual path and the skip connection.
In conclusion, rank analysis is crucial in determining the feasibility of reconstructing the input data in convolutional layers, especially when the activation function is not fully invertible. 

\section{Results and Discussions}
\label{Results}
\begin{figure}[!t]
     \centering
     \begin{subfigure}[b]{0.2\textwidth}
         \centering
         \includegraphics[width=\textwidth]{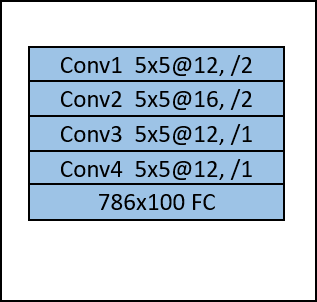}
         \caption{LeNet}
         \label{fig:Letnet}
     \end{subfigure}
     \hfill
     \begin{subfigure}[b]{0.2\textwidth}
         \centering
         \includegraphics[width=\textwidth]{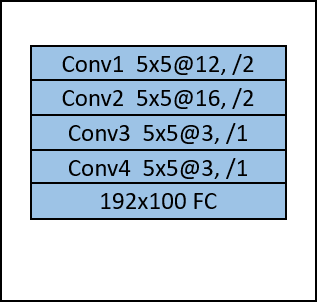}
         \caption{LeNet-O}
         \label{fig:letnet_O}
     \end{subfigure}
     \hfill
     \begin{subfigure}[b]{0.2\textwidth}
         \centering
         \includegraphics[width=\textwidth]{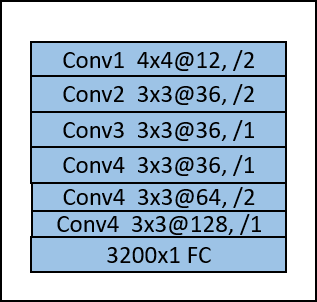}
         \caption{CNN6}
         \label{fig:CNN6}
     \end{subfigure}
          \hfill
     \begin{subfigure}[b]{0.2\textwidth}
         \centering
         \includegraphics[width=\textwidth]{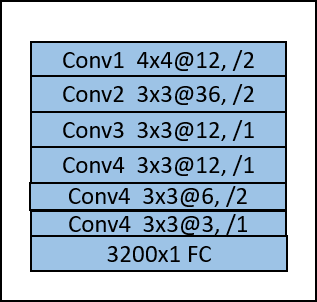}
         \caption{CNN6-O}
         \label{fig:CNN6-O}
     \end{subfigure}
        \caption{The four network architectures used for evaluation: (a) LeNet, as defined in DLG, serves as a baseline model. (b) CNN6, as defined in R-GAP, represents another standard architecture. (c) LeNet-O, a modified version of LeNet, demonstrates the effectiveness of our method in using fewer parameters for image reconstruction. (d) CNN6-O, a modified version of CNN6, reduces the number of filters in the last layer from 128 to 3, highlighting our method's ability to reconstruct images with minimal parameters. In the notation for each layer, such as ``Conv1 5x5@12, /2'', ``5x5'' specifies the kernel size, ``12'' represents the number of filters, and ``/2'' denotes the stride value.}
        \label{fig:Different arrangements of layers}
\end{figure}

\textbf{Experimential Setup:} To illustrate the effectiveness of our R-CONV method, we compared it with existing methods DLG \citep{zhu2019deep} and R-GAP \citep{zhu2020r} using three different datasets: CIFAR-100, CIFAR-10, and MNIST. We employed four different CNN architectures to evaluate our method, as shown in Figure \ref{fig:Different arrangements of layers}. These include LeNet, CNN6, and their optimized versions LeNet-O and CNN6-O (having a comparatively very small number of parameters). To compute the gradient we pass an input from the model and then compute gradients during the backward pass.  We used three performance metrics to evaluate our proposed method's effectiveness: mean squared error (MSE), peak signal-to-noise ratio (PSNR), and reconstruction time. Our experiments were conducted on a 12th Gen Intel(R) i7 2.10 GHz with 32 GB RAM, demonstrating that our method significantly improves reconstruction time compared to DLG and R-GAP. All implementation was carried out in the PyTorch framework. 

\textbf{Quantative Results:} Our novel approach, R-CONV, successfully extends existing analytical gradient attacks and provides an efficient method for reconstructing inputs of convolutional layers with minimal parameters. The experimental results demonstrate that our approach not only reconstructs images more quickly than both DLG and R-GAP but also produces higher-quality images. Our method is effective across a wide range of activation functions that other approaches fail to handle. We used LeNet and LeNet-O to compare DLG and R-CONV and CNN6 and CNN6-O for comparing R-GAP and R-CONV, respectively. Table \ref{tab:quant_results} summarizes the quantitative results of R-CONV in terms of average MSE, PSNR, and reconstruction time. Also, the table provides a comparative analysis of our proposed R-CONV method with existing methods, i.e., DLG, and R-GAP. The table demonstrates the effectiveness of our method in using fewer parameters for image reconstruction. Notably, when replacing the Sigmoid function with ReLU, DLG's reconstruction quality deteriorates significantly, whereas R-CONV maintains high quality. Moreover, the table demonstrates how much faster R-CONV is compared to R-GAP. This is because our method leverages the full set of gradient constraints to provide a more reliable and accurate reconstruction, overcoming the limitations of previous approaches (that rely on weight constraints, which require constructing the output of each layer which is not feasible in the presence of non-invertible activation functions).

%This is because our method relies on constructing the gradient w.r.t to output rather than the output itself. It involves propagating the gradient w.r.t the input of the subsequent layer through the activation function, requiring the computation of the derivative of the activation function instead of its initial input.

\begin{table}[!t]
\footnotesize
\centering
\caption{Comparison of the proposed R-CONV method with state-of-the-art analytical (R-GAP) and optimization-based (DLG) methods in terms of average MSE, PSNR, and reconstruction time. \textit{Our proposed method outperforms these state-of-the-art methods in all the considered metrics.}}
\label{tab:quant_results}

\begin{tabular}{cccc}
\toprule
 Method & MSE                                            & PSNR (dB) & Time (s) \\ \midrule 
\multicolumn{4}{c}{Average Computed for Images Depicted in Figure \ref{fig:Reconstruction_Quality_CIFAR}.}  \\ \midrule
 R-CONV  & $2.2 \times 10^{-7} \pm 3.64 \times 10^{-9} $ & $114.68\pm 5.5$ & $6.33 \pm 2.3$   \\
 DLG     & $0.0933\pm 0.05$                                       & $62.62 \pm 8.5$  & $60.66 \pm 5.58$  \\ \midrule \midrule
\multicolumn{4}{c}{Average Computed for Images Depicted in Figure \ref{Reconstruction_Quality_CIFAR}.} \\ \midrule
R-CONV  & $2.88 \times 10^{-9} \pm 2.44 \times 10^{-10} $& $150.12 \pm 4.5$ & $2.494 \pm 1.66$  \\
R-GAP   & $0.0056 \pm 0.008$                                       & $76.73 \pm 6.88$  & $232.45\pm 12.44$ \\ \bottomrule
\end{tabular}
\end{table}

\begin{figure}[!ht]
     \centering
     \begin{subfigure}[b]{0.45\textwidth}
         \centering
         \includegraphics[width=\textwidth]{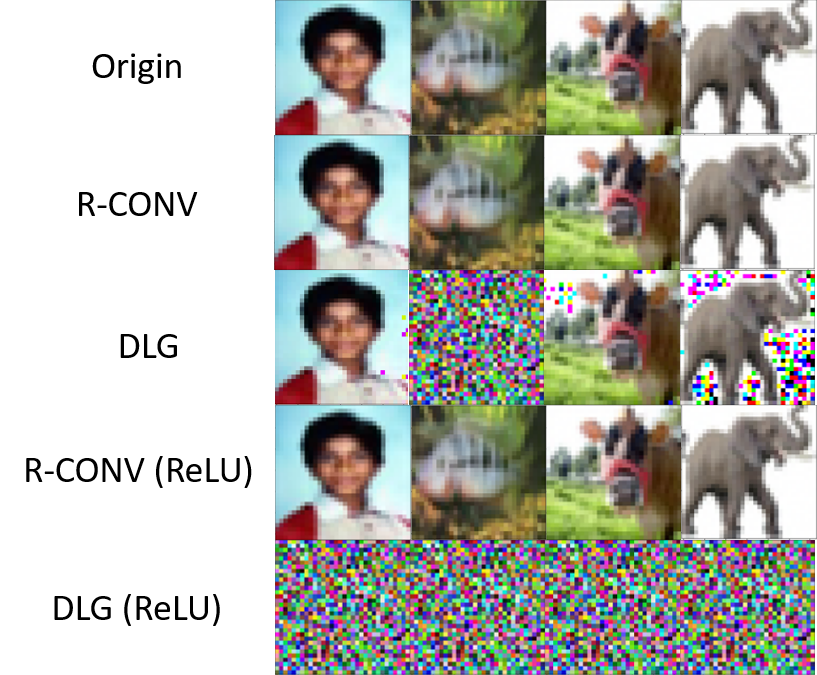}
         \caption{CIFAR100}
         \label{fig:Cifar10}
     \end{subfigure}
     \hfill
     \begin{subfigure}[b]{0.45\textwidth}
         \centering
         \includegraphics[width=\textwidth]{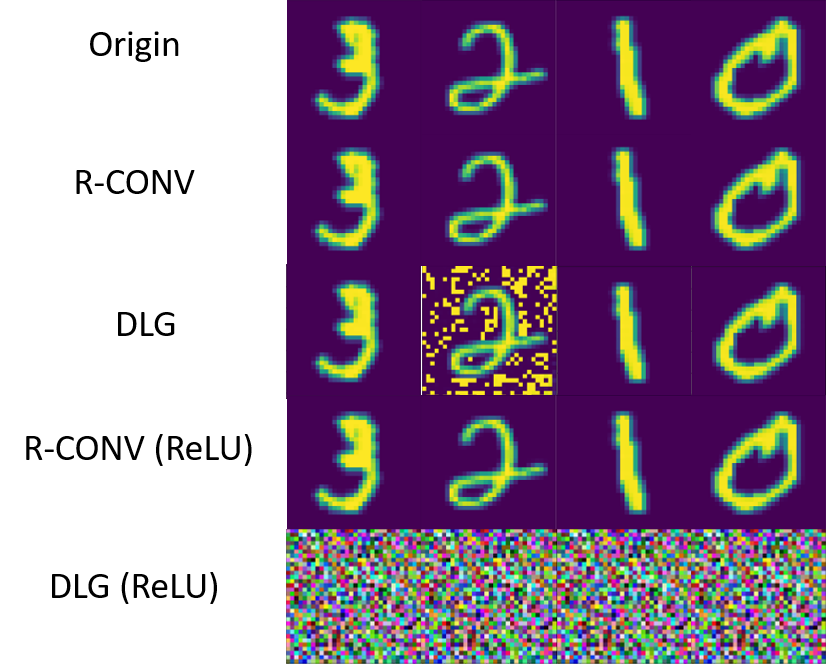}
         \caption{MNIST}
         \label{fig:MNIST}
     \end{subfigure}

        \caption{Reconstruction quality by R-CONV and DLG. Subfigure (a) shows reconstruction results using images  CIFAR-100, and (b) depicts reconstruction results using images from MNIST. The first row represents the original images, the second row shows reconstructions by R-CONV using LeNet-O, the third row shows reconstructions by DLG using LeNet, the fourth row shows reconstructions by R-CONV with ReLU activation, and the fifth row shows reconstructions by DLG with ReLU activation. \textit{The results indicate that R-CONV produces higher-quality reconstructions than DLG, and when using ReLU activation, DLG's reconstruction quality deteriorates significantly, whereas R-CONV maintains superior quality even with fewer parameters}.}
        \label{fig:Reconstruction_Quality_CIFAR}
\end{figure}

\begin{figure}[!ht]
     \centering
         \includegraphics[width=\textwidth]{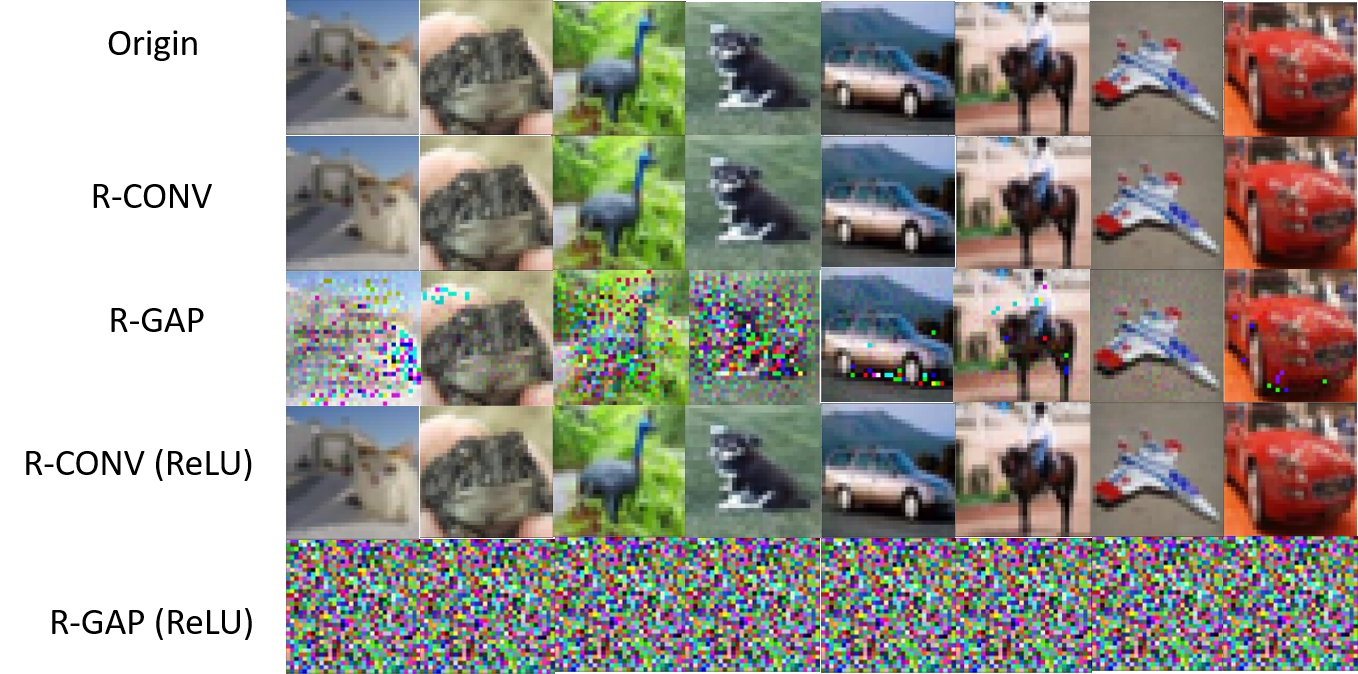}
        \caption{Reconstruction quality by R-CONV and R-GAP. This figure presents the reconstruction results of images from CIFAR-10, comparing R-CONV and R-GAP. The first row shows the original images, the second row displays reconstructions by R-CONV using CNN6-O, the third row shows reconstructions by R-GAP using CNN6, the fourth row presents reconstructions by R-CONV with ReLU activation, and the fifth row shows reconstructions by R-GAP with ReLU activation. \textit{The comparison underscores the superior performance of R-CONV, which maintains high-quality reconstructions even with ReLU activation, unlike R-GAP.}}
        \label{Reconstruction_Quality_CIFAR}
\end{figure}

\textbf{Qualitative Results:} Figure \ref{fig:Reconstruction_Quality_CIFAR} illustrates the reconstruction quality of images from CIFAR-100 and MNIST, showing superior quality of images reconstructed using our approach. Our method is effective across a wide range of activation functions where other approaches fail to reconstruct good-quality images. Notably, when replacing the Sigmoid function with ReLU, DLG's reconstruction quality deteriorates significantly, whereas R-CONV maintains high quality. For comparison with R-GAP, we used the CNN6 and CNN6-O architectures. We evaluated the performance on CIFAR-10. Figure \ref{Reconstruction_Quality_CIFAR} shows the reconstruction results of images from CIFAR-10, comparing R-CONV and R-GAP. It shows how R-CONV outperforms R-GAP in terms of reconstruction quality and its applicability to non-fully invertible activation functions, unlike R-GAP. Moreover, the results show that our method can reconstruct images with fewer parameters, as demonstrated with LeNet-O, and CNN6-O. 

% \textcolor{blue}{Illustrating reconstruction quality of different images generated using R-CONV and DLG. Figure \ref{fig:Cifar10} shows CIFAR-100 images and Figure \ref{fig:MNIST} illustrates MNIST images. The rows represent, First: original images; Second: R-CONV with LeNet-O; Third: DLG with LeNet; Foruth: R-CONV with ReLU activation; and Fifth: DLG with ReLU activation. The results indicate that R-CONV consistently achieves higher-quality reconstructions than DLG, with DLG's performance notably deteriorating with ReLU activation, while R-CONV maintains superior quality even with fewer parameters.}

\textbf{Limitations:} Although our proposed R-CONV method has achieved promising results in reconstructing good-quality images, its limitations should be acknowledged. For instance, R-CONV is primarily used to reconstruct a single training sample. Using composite activation functions (e.g., GELU, SiLU, Swish) will limit R-CONV's effectiveness in exploiting gradient constraints, as their derivative can't be expressed in terms of the output.

% \begin{table}
%   \caption{Performance metrics for image reconstruction using R-CONV, and DLG}
%   \label{table1}
%   \centering
%   \begin{tabular}{llll}
%     \toprule

%      AVERAGE & MSE     & PSNR(dB)    & Time (S) \\
%     \midrule
%     R-CONV &  2.2 x10^-^7& $114.68$ & $5.68$     \\
%    % R-GAP& $0.0933$ & $58.42$ &$220.45$      \\
%     DLG & $0.0056$&$70.62$&$62.66$  \\
%     \bottomrule
%   \end{tabular}
%   \end{table}
%   \begin{table}
%   \caption{Performance metrics for image reconstruction using R-CONV, and R-GAP}
%   \label{table2}
%   \centering
%   \begin{tabular}{llll}
%     \toprule

%      AVERAGE & MSE     & PSNR(dB)    & Time (S) \\
%     \midrule
%     R-CONV &  2.88 x10^-^9& $150.12$ & $2.494$     \\
%    R-GAP& $0.0933$ & $58.42$ &$220.45$      \\
%     \bottomrule
%   \end{tabular}
% \end{table}

\section{Conclusions}
In this paper, we focused on advancing the understanding and effectiveness of gradient inversion attacks on shared model updates in the federated learning paradigm. Recognizing the importance of understanding information leakage mechanisms, we proposed R-CONV, an advanced analytical method that efficiently reconstructs training data from the gradients of convolutional layers. Our findings emphasize the critical role of gradient constraints, which can be more revealing than weight constraints in certain layers, an aspect previously overlooked.
In addition, we conducted a detailed rank analysis and investigated the influence of various convolutional layer parameters. Our method outperforms existing techniques by achieving better data reconstruction, reducing reconstruction time, and being applicable to a broader range of activation functions.

\bibliographystyle{plainnat}
\bibliography{ref}

\end{document}